\documentclass[letterpaper,10pt,conference]{ieeeconf}
\IEEEoverridecommandlockouts
\usepackage[T1]{fontenc}
\usepackage{amsmath}
\usepackage{amssymb}
\usepackage{algorithm}
\usepackage{algorithmic}
\usepackage{array}
\usepackage{booktabs}
\usepackage[nocompress]{cite}
\usepackage{graphicx}
\usepackage{tabularx}
\usepackage{url}
\usepackage[hidelinks]{hyperref}

\newcolumntype{Y}{>{\centering\arraybackslash}X}
\newcommand{\systemname}{\textsc{KnowDemo}}
\title{\LARGE \bf
KnowDemo: Knowledge-Guided Robot Demonstration Generation
from Human Videos
}

\author{Zhiyuan Gao$^{1}$, Yanxiang Zhan$^{1}$, Mohammad Khoshnazar$^{1}$,
Jeroen Sch\"afer$^{1}$, and Michael Beetz$^{1,2}$\\[3pt]
\normalsize $^{1}$University of Bremen\\
\normalsize $^{2}$Robotics Institute Germany (RIG)%
\thanks{Email addresses, in author order:
\textnormal{gao@uni-bremen.de},
\textnormal{yanxiang@uni-bremen.de},
\textnormal{khoshnam@uni-bremen.de},
\textnormal{jeroen.schaefer@uni-bremen.de}, and
\textnormal{beetz@informatik.uni-bremen.de}.}}

\begin{document}
\bstctlcite{IEEErefcontrol}

\maketitle
\raggedbottom
\thispagestyle{empty}
\pagestyle{empty}

\begin{abstract}
Learning robot manipulation policies typically requires substantial
demonstration data, which are costly to collect on real robots. Recent methods generate robot demonstrations from human videos by
adapting recovered motion and validating the resulting trajectories in
simulation. However, methods centered on motion-reference
adaptation can limit behavioral diversity by retaining the demonstrated
contact strategies and subtask orders, while insufficient
understanding of task requirements and scene relations can reduce
demonstration generation efficiency by generating invalid candidates.
To address these limitations, we propose \systemname{}, a framework that
uses structured manipulation knowledge from human videos to generate
diverse robot demonstrations for a target workspace.
To distinguish task requirements from demonstration-specific choices, we
develop a knowledge extraction and reasoning module based on a
vision-language model (VLM) that
associates object and action descriptions with inferred task conditions, demonstration references, and
permissible execution variations. To translate this knowledge into
executable demonstrations, we resolve the descriptions against target-scene
entities and geometry to guide candidate generation and screening before
motion planning and simulation. The resulting demonstrations exhibit multimodal behavior through
alternative contact strategies and valid subtask orders, with structured
execution labels.
Experiments demonstrate additional verified execution modes beyond a
reference-only configuration and improved candidate planning success
through task-guided grasp sampling.
To validate the generated data for policy learning, we fine-tune the
pretrained $\pi_{0.5}$ model on simulation data, achieving sim-to-real
transfer across three tasks.
Project page: {\urlstyle{same}\url{https://zhiyuan-gao.github.io/knowdemo/}}.
\end{abstract}

\begin{figure*}[t]
  \centering
  \includegraphics[width=\textwidth]{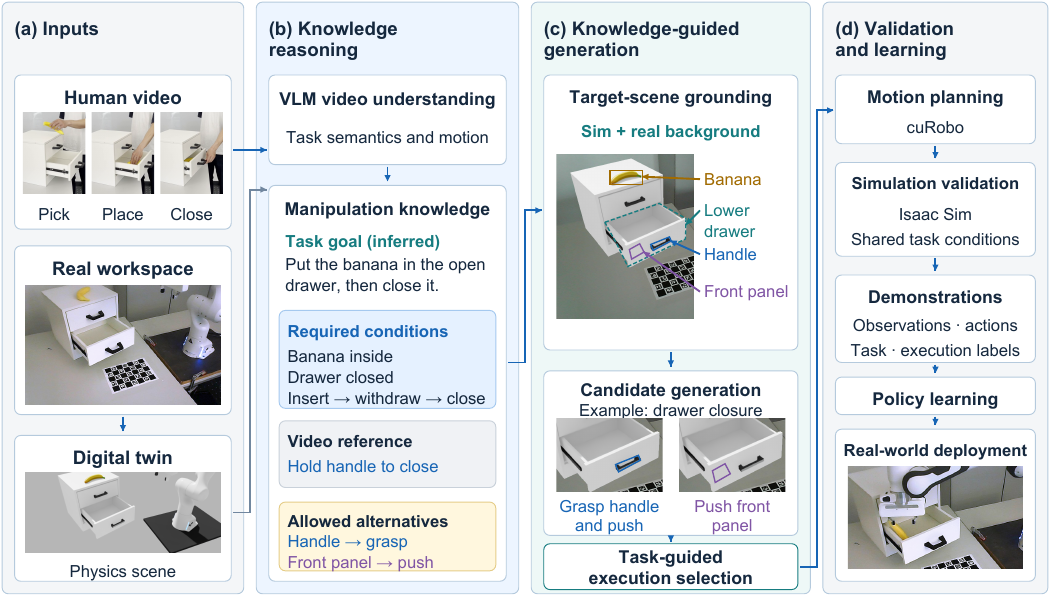}
  \caption{Overview of \systemname{}. (a) Pick, place, and close frames from a human demonstration, the target workspace, and its digital twin. (b) A drawer-closure example links an inferred task goal to required conditions, the video reference, and allowed alternatives. (c) Colors link alternatives to grounded regions and candidate contacts. Rendered robots and objects are composited onto the real background to form policy observations. (d) Planning and simulation validate candidates, yielding demonstrations with shared task instructions and execution labels for $\pi_{0.5}$ fine-tuning and real-world deployment, illustrated by banana placement during drawer storage.}
  \label{fig:system}
\end{figure*}

\section{Introduction}
\label{sec:introduction}

Robot demonstration datasets support learning manipulation policies across
tasks and environments~\cite{oxe2024rtx,khazatsky2024droid}.
However, real-robot demonstration collection remains
costly~\cite{mandlekar2023mimicgen,xue2025demogen}.
Human videos offer accessible task knowledge and motion evidence for generating
robot demonstrations~\cite{jain2024vid2robot,shi2025zeromimic,dan2025xsim}.
Turning these videos into robot training data requires generating observations
and actions for the target robot and its workspace.
Video-to-robot approaches extract task plans or recover motion to guide
execution~\cite{wang2024seedo,heppert2024ditto,mu2026deximit}.
Demonstration-generation methods expand datasets by adapting reference
motions to new object poses and
geometries~\cite{mandlekar2023mimicgen,lin2025cpgen}.
Task and scene knowledge further guide execution by grounding abstract
action descriptions in the current environment~\cite{beetz2025cram} and
incorporating semantic and geometric constraints into motion
planning~\cite{garrett2025skillgen,wang2025hybridgen,lin2025cpgen}.
Simulation subsequently tests the resulting trajectories for physical
feasibility~\cite{mu2026deximit,zhai2026psi}.
These efforts extend demonstration generation across scene and object variations.

However, adapting recovered object motions to new scene configurations can
retain the demonstrated contact strategy and subtask order. The resulting
data may cover different object configurations while capturing only a limited
range of ways to complete the task. A single task can admit multimodal
behavior~\cite{chi2023diffusion,shafiullah2022bet,jia2024d3il}.
For example, the person in the video may close a drawer by holding its
handle, whereas a robot could achieve the same goal by pushing the front panel.
Generating such alternative behaviors requires distinguishing essential task
conditions from demonstration-specific choices.

Candidate generation must also account for subsequent subtasks and
target-scene geometry. A grasp that permits object pickup may obstruct later
placement or gripper withdrawal. Ignoring these requirements can lead to
failed planning attempts. Grounding task knowledge in the target scene allows
these requirements to guide candidate generation before motion planning and
simulation.

To address these challenges, we propose \systemname{}, a framework that
uses structured manipulation knowledge from human videos to generate
diverse robot demonstrations for a target workspace (Fig.~\ref{fig:system}).
To distinguish task requirements from demonstration-specific choices, our
knowledge extraction and reasoning module based on a vision-language
model (VLM) associates
structured object and action descriptions with inferred task conditions,
observed contacts and motions, and permissible execution variations.
To translate this knowledge into executable demonstrations, we resolve
the descriptions against target-scene entities and geometry, using the
grounded task conditions to guide candidate sampling and screening before
motion planning and simulation. Each accepted demonstration retains the task instruction
inferred from the video and structured execution labels, allowing data to be organized
and selected by execution mode. The resulting simulation demonstrations
support policy fine-tuning for deployment in the target workspace.
We quantify behavioral diversity by comparing the number of verified
execution modes with a reference-only configuration, supported by examples
of alternative contact strategies and subtask orders from the same human
video. A shared-backend ablation measures the effect of task-guided grasp
sampling on candidate planning success. To validate the generated data
for robot learning,
we fine-tune the pretrained $\pi_{0.5}$ vision-language-action (VLA)
model~\cite{pi2025pi05} on the generated simulation demonstrations and
demonstrate sim-to-real transfer.

In summary, our main contributions are:
\begin{itemize}
  \item To generate robot training data from human videos, we introduce
  \systemname{}, a framework that uses structured manipulation knowledge
  to generate diverse demonstrations for a target workspace.
  \item To distinguish task requirements from demonstration-specific
  choices, we develop a knowledge extraction and reasoning method that
  associates task conditions, demonstration references, and permissible
  variations with structured object and action descriptions. To translate
  this knowledge into executable demonstrations, we ground object and action
  descriptions in target-scene entities and geometry to guide candidate
  sampling and screening before motion planning and simulation, retaining
  structured execution labels with the generated demonstrations.
  \item To evaluate the framework, we quantify gains in verified execution
  modes over a reference-only configuration, measure the improvement in
  candidate planning success, and validate sim-to-real transfer using
  policies fine-tuned on the generated data.
\end{itemize}

\section{Related Work}
\label{sec:related}

\subsection{Manipulation Knowledge from Human Videos}

Human videos support robot manipulation through two common approaches:
training policies that map robot observations to actions, and extracting
explicit knowledge representations of goals, object relations, and motion
references for planning. In the first approach, MimicPlay learns latent
plans from human play to guide a policy trained on robot
demonstrations~\cite{wang2023mimicplay}, while Vid2Robot conditions actions
on demonstration videos and current observations~\cite{jain2024vid2robot}.

Explicit representations expose task structure and motion for reasoning
and adaptation. SeeDo converts videos into action plans through
vision-language reasoning~\cite{wang2024seedo}; ORION represents object
states, contacts, and motion with Open-World Object Graphs~\cite{zhu2026orion};
DITTO transforms object-relative motion for the current
scene~\cite{heppert2024ditto}. DemoDiffusion combines motion references
with learned action priors, refining retargeted human motion using a
pretrained diffusion policy~\cite{park2026demodiffusion}.
Our structured knowledge separates task requirements, demonstration
references, and permissible variations to guide target-scene demonstration
generation.

\subsection{Knowledge-Guided Robot Demonstration Generation}

Task and scene knowledge guides contact selection and motion parameters.
CRAM represents objects, locations, and actions through designators;
perception and reasoning resolve their unspecified properties to
contextualize generalized plans~\cite{beetz2025cram}. KnowRob combines
symbolic knowledge, perception, geometry, and simulation for manipulation
reasoning~\cite{beetz2018knowrob}. Spatial constraints and value maps
provide geometric guidance for planning~\cite{manuelli2019kpam,huang2025rekep,huang2023voxposer}.
Task-oriented grasping connects contact selection to the intended operation:
OVAL-Grasp uses task-relevant part heatmaps to score and filter grasp
candidates~\cite{tong2025ovalgrasp},
and DemoFunGrasp conditions grasps on affordance regions and grasping
styles~\cite{mao2026demofungrasp}.

Demonstration generators make these decisions across scene configurations.
MimicGen and DemoGen adapt references to new object
poses~\cite{mandlekar2023mimicgen,xue2025demogen}; SkillMimicGen combines
skill-level generation with planning~\cite{garrett2025skillgen}.
CP-Gen preserves keypoint-trajectory constraints across
geometries~\cite{lin2025cpgen}, HybridGen uses VLM segment classification
and constraint-guided replanning~\cite{wang2025hybridgen}, and AffordGen
transfers affordance keypoints across meshes~\cite{zhang2026affordgen}.
Using human videos, DexImit, PSI, and Real2Gen combine motion synthesis
with simulation~\cite{mu2026deximit,zhai2026psi,heppert2026real2gen}.
Pegasus combines task, affordance, and constraint graphs with physical
verification~\cite{luo2026pegasus}.

We adopt CRAM's designators and resolution mechanism to link video-derived
task conditions, references, and permissible variations to action descriptions.
Target-scene resolution specifies alternative contact regions, interaction
modes, grasp strategies, and valid subtask orders. Generated demonstrations
retain these choices as execution labels under the shared task instruction.

\subsection{Scene Reconstruction for Robot Simulation}

Video2Policy, Human2Sim2Robot, and X-Sim reconstruct interactive environments
for learning from human videos~\cite{ye2025video2policy,lum2025human2sim2robot,dan2025xsim}.
Articulated-object reconstruction supplies geometry and
joints~\cite{weng2024digitaltwin}. For visual matching, SIMPLER composites
simulated assets onto real backgrounds~\cite{li2024simpler}, while SplatSim
and RoboSimGS use Gaussian splatting~\cite{qureshi2025splatsim,zhao2025robosimgs}.
Our target-workspace digital twin supplies entity references and geometry
for knowledge grounding and demonstration generation across tasks.

\section{Method}
\label{sec:method}

\systemname{} comprises four components: digital-twin construction
(Section~\ref{sec:overview}), manipulation knowledge extraction and
reasoning (Section~\ref{sec:task-abstraction}), knowledge-guided execution
generation (Section~\ref{sec:grounding}), and simulation validation and
policy learning (Section~\ref{sec:policy}). Figure~\ref{fig:system} shows
the framework.

\subsection{Problem Setup and Digital-Twin Construction}
\label{sec:overview}

Given a human video $V$, calibrated target-workspace RGB observations
$Y_R$ (optionally with depth), and a robot model $M_R$, we generate an
observation--action dataset $D_R$. Demonstrations satisfy the inferred task
requirements while allowing alternatives to the observed contacts and
subtask order.

We construct a digital twin $\mathcal E_R$ of the target workspace in
Isaac Sim, using an official Franka model, reconstructed task objects,
and an aligned collision table.
The table provides physical support and collision geometry but is excluded
from the rendered RGB foreground. Policy observations combine robot
and object renderings with a fixed-camera workspace image after masking
and inpainting movable entities~\cite{li2024simpler}.
SAM3~\cite{carion2025sam3} segments images, SAM3D
Objects~\cite{sam3d2025} reconstructs rigid objects, and
URDF-Anything+~\cite{wu2026urdfanything} reconstructs articulated task
objects, including drawer cabinets and microwave doors, with their part
geometry and joints.
Assets are scaled using known dimensions or depth, calibrated to the robot
base, and converted to USD with joint and physical properties. Collision
meshes preserve cavities and contact surfaces.

A queryable USD index~\cite{nguyen2024usd} stores entity IDs, prim paths,
poses, meshes, part hierarchies, and joint types, axes, and limits. Rendered
masks link IDs to image regions; back-projected part masks and fitted
interiors define contact patches and containment volumes for grounding.

\subsection{Manipulation Knowledge Extraction and Reasoning}
\label{sec:task-abstraction}

To distinguish task requirements from demonstration-specific choices, our
knowledge extraction and reasoning module combines VLM-based parsing of
object roles and subtask structure with recovered motion evidence.
A pretrained VLM with frozen parameters reasons over this information and
target-scene descriptions, using commonsense knowledge to infer task
conditions and permissible execution variations.

\paragraph{Video parsing}
The VLM infers a task instruction from the video and identifies subtask
boundaries from changes in manipulated objects, contacts, and object relations. Each subtask records its interval, object
roles, initial and goal relations, contact/release events, and supporting
frames. SAM3.1~\cite{sam2026sam31} tracks object masks;
SpatialTracker V2~\cite{xiao2025spatialtrackerv2} estimates depth and camera
parameters; FoundationPose++~\cite{teal2025foundationposepp} uses these
estimates and aligned SAM3D Objects meshes to track object and rigid-part
poses. Motion relative to reference object $b$ is
${}^{b}T_o(t)=({}^{C}T_b(t))^{-1}{}^{C}T_o(t)$, where $C$ is the source
camera frame and ${}^{A}T_B$ maps $B$ coordinates to $A$.
Without known source dimensions, translations are normalized by source
extent and scaled to the corresponding target dimension. Motion intervals
are linked to their subtasks.

\paragraph{Structured knowledge}
We adopt CRAM's object, location, and action
designators~\cite{beetz2025cram}. Object designators describe properties
and parts; location designators specify spatial relations and pose
requirements; action designators reference these descriptions and contain
execution parameters to be resolved in the target scene.

The VLM organizes the linked knowledge as
$\mathcal K=(\mathcal O,\mathcal S,\mathcal C,\mathcal R,\mathcal A)$:
$\mathcal O$ contains object and part descriptions, $\mathcal S$ contains
subtasks and their dependencies, $\mathcal C$ specifies required conditions,
$\mathcal R$ stores contact and motion references, and $\mathcal A$ specifies
permissible variations. Records link to the relevant designators and
subtasks; target-scene resolution instantiates alternatives using available
parts, geometry, and execution operators.
Conditions specify predicates, role arguments, and
whether they apply before, during, or after a subtask; variations identify candidate contact
parts and supported interaction modes. Predicates cover containment,
relative placement, orientation, joint state, release, and gripper clearance.

Observed order is recorded in $\mathcal R$, while precedence edges in
$\mathcal S$ are inferred when one subtask produces a state required by
another. Video observations carry evidence links and are distinguished
from commonsense deductions. Structural checks verify role references,
acyclic subtask dependencies, and supported predicates and modes. When ambiguities or
inconsistencies arise, the VLM re-examines relevant clips and target-scene
views; unresolved critical fields stop generation.

\subsection{Knowledge-Guided Execution Generation in the Target Workspace}
\label{sec:grounding}

We use CRAM's designator resolution mechanism to instantiate descriptions
in the digital twin. Our resolution procedures combine
VLM-based semantic matching, USD queries, and geometric sampling
(Algorithm~\ref{alg:generation}).

\begin{figure}[!t]
  \centering
  \includegraphics[width=0.82\columnwidth]{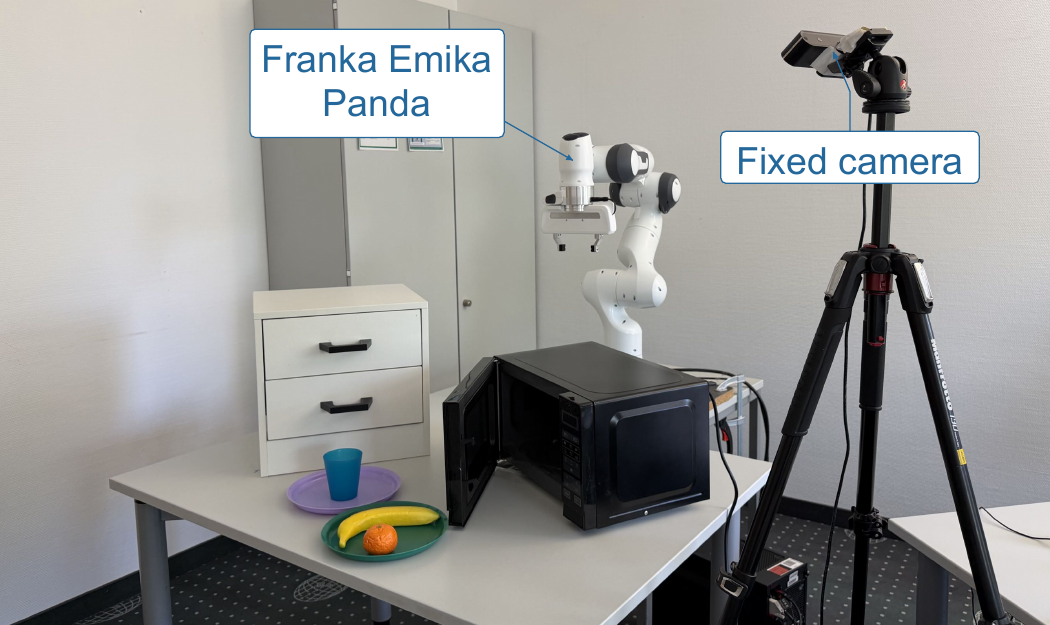}
  \caption{Real-robot workspace with a Franka arm, parallel-jaw gripper, and fixed external camera.}
  \label{fig:robot-workspace}
\end{figure}

\paragraph{Entity and region grounding}
The VLM uses the digital twin's entity and part descriptions, spatial
relations, joint states, geometric queries, and rendered views with ID
overlays to match object references to scene entities. Each reference $r$
is bound to an entity and region, $\beta(r)=(e_R,\rho_R)$. The scene index
verifies entity existence, part membership, and joint compatibility. The
location resolver converts spatial relations and pose requirements into
sampling conditions; concrete poses are sampled after each layout update.

\paragraph{Execution selection}
Given the inferred task conditions and permissible variations, the VLM
uses digital-twin information to select candidate contact parts, interaction
modes, and grasp strategies supported by the system. For grasping, it also
selects a coarse approach direction compatible with subsequent placement,
release, and withdrawal. Each alternative is $\xi=(p,m,h)$, with contact
part/region $p$, interaction mode $m$, and grasp strategy $h$; surface
pushing uses $h=\varnothing$. For grasping alternatives, an associated
approach-direction rule constrains geometric grasp sampling. Supported
strategies include rim grasping,
body grasping, and handle grasping, corresponding to contacts on the rim,
outer body, and handle. The available parts determine which strategies
apply to an object. The VLM retains multiple task-compatible alternatives.

A task-constraint adapter maps modes to grasp-and-transport,
placement/release, grasped-articulation, or surface-pushing operators.
Each operator provides geometric inputs, gripper states,
preconditions/effects, and a waypoint constructor. Task dependencies and
operator preconditions determine admissible subtask orders. Typed predicates
check preconditions, propagated effects, contact regions, and goal
conflicts before planning; continuous conditions remain attached to their
subtasks for geometric and physical validation.

\paragraph{Geometric sampling and motion construction}
The adapter obtains contact regions from bound parts and resolves
admissible approach directions from their geometry. A grasp strategy may
admit multiple approach directions.

We represent each admissible approach direction using the approach-cone
formulation in CAPGrasp~\cite{weng2024capgrasp}. The adapter resolves a unit direction
$\mathbf d$ in the robot-base frame from gravity, a part frame, or a
surface normal. With angular tolerance $\alpha$, a grasp $g$ must satisfy
\begin{equation}
 \mathbf u(g)^{\mathsf T}\mathbf d\geq\cos\alpha,
 \label{eq:grasp-approach-cone}
\end{equation}
where $\mathbf u(g)$ is the unit pre-grasp-to-contact direction. The
tolerance is specified by the geometric rule for that approach. Top-down
and side describe approach directions; rim, body, and handle describe
the grasp strategies whose contacts are constrained by the bound regions.

After each layout update, the Isaac Sim Grasping SDG antipodal
sampler~\cite{nvidia2025graspingsdg} generates concrete grasp poses from
object meshes, with opposing surface contacts within the gripper aperture.
The adapter retains poses satisfying the selected strategy's grounded
contact-region, approach-direction, and clearance constraints. Rotation
about the approach axis remains available. Placement sampling accounts for
destination boundaries, object footprint, support height, and orientation.

For compatible interaction modes, we express the recovered object motion
relative to the task-relevant reference object or part. We map this motion
into the corresponding target frame, scale translations to target dimensions,
and adapt it to the sampled placement pose. For candidate $a$, its object waypoints
and sampled grasp $g_a$ determine the end-effector targets:
\begin{equation}
 {}^{W_R}T_e(t;a)={}^{W_R}T_o(t;a)\,{}^{o}T_e(g_a),
 \label{eq:object-to-end-effector}
\end{equation}
where $W_R$ is the robot-base frame and ${}^{o}T_e(g_a)$ is the grasp's
object-relative gripper pose. Alternative interaction modes construct waypoints
from sampled contacts, goals, and operator geometry. Each candidate includes
release and withdrawal.

\paragraph{Drawer example}
In Fig.~\ref{fig:system}(b)--(c), the required outcome is a closed drawer
with the banana inside. The human's handle grasp is a reference; resolution
yields handle-grasp and panel-push alternatives. Handle-based motion adapts
the reference to the target joint geometry. Panel pushing uses the sampled
contact, panel normal, joint axis, and drawer displacement. Both follow
insertion and gripper withdrawal and satisfy the same containment and
joint-state conditions (Fig.~\ref{fig:execution-diversity}(a)).

\paragraph{Planning}
cuRobo~\cite{sundaralingam2023curobo} receives the joint state, collision
world, carried-object geometry, and targets. Free-space segments use motion
generation; interaction waypoints use sequential inverse kinematics
initialized from the preceding solution. Interpolated paths are checked
for joint limits, collisions, and orientation/region conditions, with
intended contacts specified per subtask.

\begin{algorithm}[t]
\caption{Knowledge-guided demonstration generation}
\label{alg:generation}
\begin{algorithmic}[1]
\REQUIRE Knowledge $\mathcal K$, indexed digital twin $\mathcal E_R$, attempt budget $B$
\STATE Resolve designators to bindings $\beta$, sampling conditions, and alternatives $\xi=(p,m,h)$ with grasp approach rules
\STATE Enumerate admissible alternative/order combinations; initialize $D_R\leftarrow\emptyset$
\FOR{$i=1$ to $B$}
 \STATE Select the next combination in round-robin order; sample $x_0$ and update grounded geometry
 \STATE \textbf{if} symbolic conditions fail \textbf{then continue}
 \STATE Sample contacts, grasps, and goals under grounded constraints
 \STATE \textbf{if} geometric sampling fails \textbf{then continue}
 \STATE Construct waypoints from compatible references or operator geometry; plan the complete execution
 \STATE \textbf{if} planning fails \textbf{then continue}
 \STATE Reset to $x_0$; execute and monitor subtask conditions
 \STATE Verify execution labels against the rollout
 \IF{all subtask conditions, final goals, and label checks pass}
  \STATE Store the task instruction, verified labels, observations, and actions in $D_R$
 \ENDIF
\ENDFOR
\RETURN $D_R$
\end{algorithmic}
\end{algorithm}

\subsection{Simulation Validation and Policy Learning}
\label{sec:data-generation}
\label{sec:policy}

Isaac Sim executes joint/gripper commands and checks subtask predicates using
relative gripper--object motion, placement stability, joint states, and
destination geometry. Acceptance requires all subtask conditions and final
goals, including release and withdrawal; failures are logged by subtask.
Accepted records synchronize RGB, proprioception, actions, knowledge,
and bindings. They retain the shared task instruction and execution labels:
target entities, contact parts, interaction modes, grasp strategies, and
subtask order. Execution traces and contact checks verify labels initialized
from candidate choices. An execution mode is defined by a contact strategy
and a valid subtask order. Continuous contact positions and target-plate
choices are not counted as separate modes. Labels support selection by
execution mode and optional conditioning of policy training. Training actions comprise end-effector pose increments
computed from consecutive executed poses and the corresponding binary
gripper commands. Policy fine-tuning and real deployment use the same
action representation.
Section~\ref{sec:real-evaluation} specifies the learning and deployment settings.

\section{Experiments}
\label{sec:setup}
\label{sec:results}

We evaluate three aspects of the framework: behavioral diversity from the
same human video, candidate planning success through task-guided grasp
sampling, and the use of generated data for policy fine-tuning and
sim-to-real transfer.

\begin{figure}[!t]
  \centering
  \begin{minipage}{0.90\columnwidth}
  \input{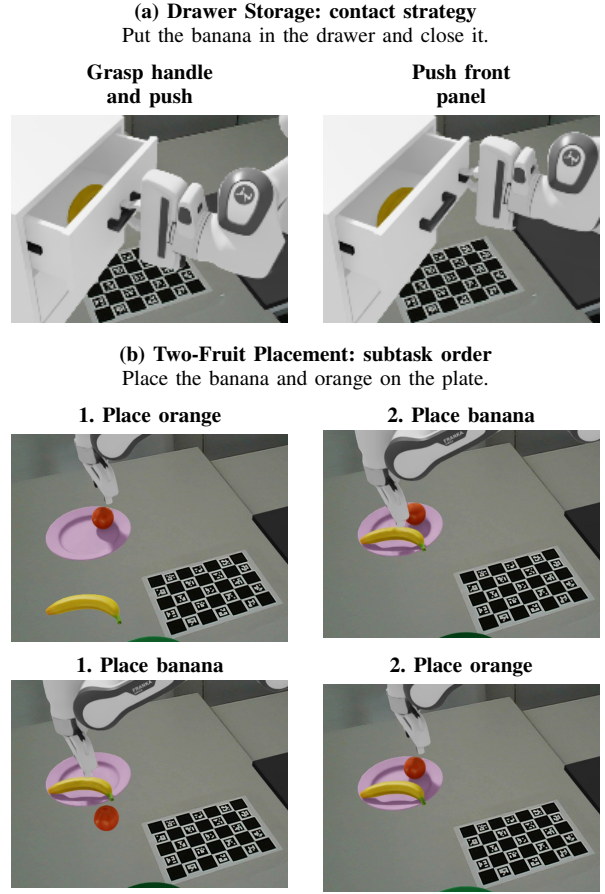}
  \end{minipage}
  \caption{Successful simulation executions under shared task instructions. For each task, alternatives use the same reference video and initial state. (a) Drawer closure by grasping the handle and pushing, or by pushing the front panel. (b) Both fruit-placement orders on the same plate, shown after each placement.}
  \label{fig:execution-diversity}
\end{figure}

\subsection{Experimental Setup}

\paragraph{Platform and inputs}
We use a Franka arm with a parallel-jaw gripper, a fixed external RGB
camera, and the digital twin described in Section~\ref{sec:overview}
(Fig.~\ref{fig:robot-workspace}). We run the pretrained
Qwen3.8-27B~\cite{qwen2026qwen38} with frozen parameters on a single NVIDIA
A100 GPU. Videos are sampled at 2 FPS with timestamps;
ambiguous contacts or subtask boundaries trigger VLM re-analysis of more
densely sampled clips.
Each task uses one reference video, recovered motion evidence, and target
assets. Layouts vary object poses and include distractors.

\paragraph{Tasks}
\textit{Drawer Storage} opens a drawer, inserts a banana, withdraws the
gripper from the interior, and closes the drawer with the banana inside;
closure permits handle grasping or panel pushing. \textit{Two-Fruit
Placement} places an orange and a banana on a plate in either order,
leaving both stable within their placement regions. \textit{Close Microwave}
requires full door closure and gripper withdrawal, allowing different
pushing contacts. Real-policy evaluation uses these three tasks, with
Banana Drawer Storage starting from an already open drawer.

\begin{figure}[!t]
\centering
\includegraphics[width=\columnwidth]{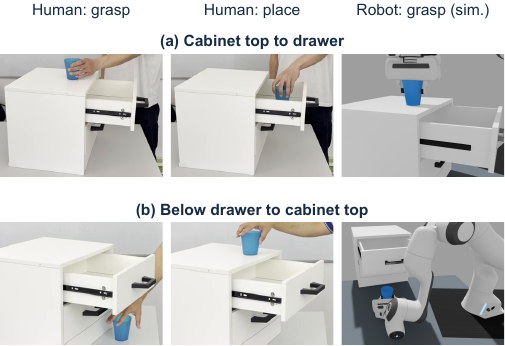}
\caption{Human grasp and placement references with corresponding robot grasps in simulation: (a) cabinet top to drawer, using a top-down robot approach; (b) beneath an extended drawer to cabinet top, pairing a palm-down human grasp with a side approach by the robot. Robot grasps accommodate placement and withdrawal in (a), and access beneath the drawer in (b).}
\label{fig:grasp-task-setups}
\end{figure}

Grasp-sampling evaluation uses the two tasks in Fig.~\ref{fig:grasp-task-setups}:
\textit{Cup-in-Drawer} moves a cup from the cabinet top into an open drawer,
and \textit{Cup-to-Cabinet} moves a cup from beneath an extended drawer
onto the cabinet top.
Both require stability after release and gripper withdrawal. Success
criteria are fixed independently of inferred knowledge.

\paragraph{Generation backend}
The backend combines video-derived motion references, geometric grasp
sampling, and planning, following the general workflow of
Real2Gen~\cite{heppert2026real2gen}, with task adapters for articulated
motion and execution of multiple subtasks. The grasp-sampling comparison changes
grasp-strategy selection within this shared backend, as detailed below.

\subsection{Behavioral Diversity of Generated Demonstrations}
\label{sec:diversity-evaluation}

Table~\ref{tab:verified-modes} reports execution modes verified by complete
successful demonstrations. \textit{Reference-only} retains the demonstrated
contact strategy and subtask order using the same backend for target-scene
adaptation. Modes follow the definition in Section~\ref{sec:policy}.

\begin{table}[!t]
\centering
\caption{Verified execution modes.}
\label{tab:verified-modes}
\small
\begin{tabularx}{\columnwidth}{@{}l Y Y@{}}
\toprule
Task & Reference-only & \systemname{} \\
\midrule
Drawer Storage & 1 & 2 \\
Two-Fruit Placement & 1 & 2 \\
Close Microwave$^{\ast}$ & 1 & 1 \\
\bottomrule
\end{tabularx}
\par\smallskip
\begin{minipage}{\columnwidth}
\footnotesize $^{\ast}$\textbf{Close Microwave:} both use one pushing mode;
\systemname{} varies contact positions within that mode.
\end{minipage}
\end{table}

Figure~\ref{fig:execution-diversity} shows successful alternatives from
matched initial states. Drawer Storage uses handle grasping or panel pushing
for closure after banana insertion and gripper withdrawal, preserving
containment and the final closed state (Fig.~\ref{fig:execution-diversity}(a)).

For Two-Fruit Placement, \systemname{} varies placement order and, when
task conditions permit, target plate; both are retained as execution labels.
Figure~\ref{fig:execution-diversity}(b) shows banana-first and orange-first
sequences using the same plate, with frames after each placement.

For Close Microwave, the generated demonstrations use different door
contact positions within the same pushing mode, while preserving the task
instruction and requiring no additional human videos.

\begin{table}[!t]
\centering
\caption{Grasp-sampling ablation.}
\label{tab:generation-efficiency}
\small
\begin{tabularx}{\columnwidth}{@{}l Y Y@{}}
\toprule
Task & \shortstack{Without task\\guidance} & \systemname{} \\
\midrule
Cup-in-Drawer & 29/50 & 50/50 \\
Cup-to-Cabinet & 23/50 & 45/50 \\
\bottomrule
\end{tabularx}
\par\smallskip
{\footnotesize Entries report successful complete-task plans / sampled candidates.}
\end{table}

\begin{figure*}[!t]
  \centering
  \begin{minipage}{\textwidth}
  \input{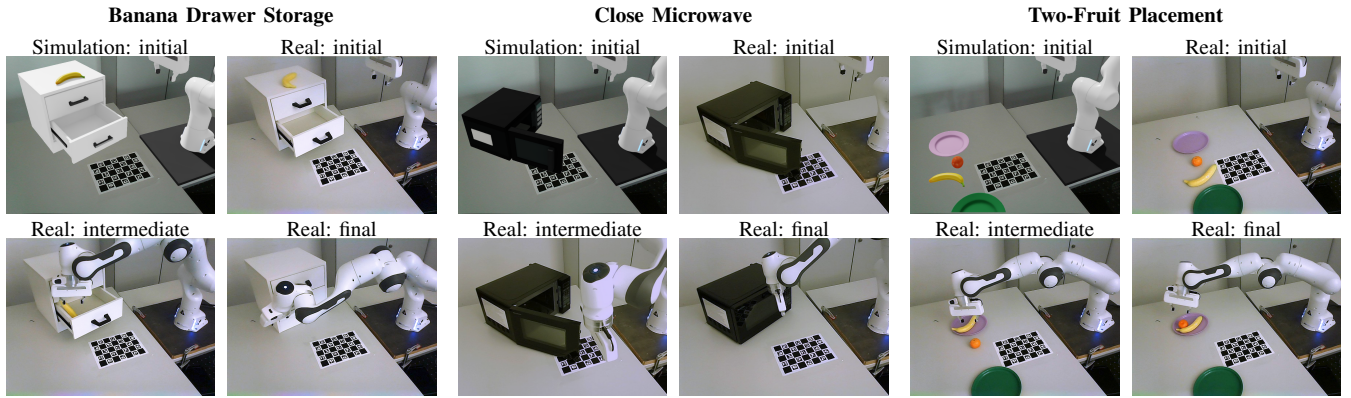}
  \end{minipage}
  \caption{Simulation observations and real-robot executions. For each task, the upper-left image combines rendered robots and task objects with a real background. The other three images show initial, intermediate, and final frames of autonomous real-robot executions using $\pi_{0.5}$ fine-tuned on generated simulation demonstrations.}
  \label{fig:platform-rollouts}
\end{figure*}

\subsection{Task-Guided Grasp Sampling}

Human grasp strategies need not transfer directly to a robot gripper.
Figure~\ref{fig:grasp-task-setups} pairs grasp and placement frames from
each reference video. In (a), the cup moves from the cabinet top into a
drawer, where placement and gripper withdrawal constrain grasp choice.
The human reorients the cup within the hand before insertion, so the grasp
and placement frames show different hand--object configurations.
In (b), the human uses a palm-down grasp to move the cup from beneath an
extended drawer to the cabinet top; the overhang constrains robot approach
and extraction. The framework preserves the task goals while selecting
robot contact regions and approach directions from target geometry.
Planning and simulation check the complete execution.

Task-oriented grasping connects contact selection to operation
requirements~\cite{tong2025ovalgrasp}. Our quantitative ablation uses
Cup-in-Drawer and Cup-to-Cabinet to measure whether task-guided strategy
selection increases complete-task planning success under a fixed budget.

\paragraph{Comparison}
We isolate task-guided grasp selection within the same generation backend.
\textit{Without task guidance}, a grasp strategy and its associated
approach rule are sampled from the object's applicable set independently
of task-specific approach and placement requirements.
\systemname{} selects strategies using task requirements and target-scene
information. For the handleless cups, both configurations use the rim- and
body-grasp strategies from Section~\ref{sec:grounding}. The adapter maps each
strategy to contact-region and approach-direction constraints.
\systemname{} selects a top-down approach for Cup-in-Drawer to accommodate
placement and withdrawal inside the drawer, and a side approach for
Cup-to-Cabinet to accommodate access beneath the extended drawer.
Both configurations use Grasping SDG with the same scene assets, grounded
geometry, placement goals, execution operators, and per-candidate planning
budget; task conditions and geometric checks remain active during planning
and simulation validation.
Initial positions vary within a small neighborhood of each setup, with
paired states across methods. Each method submits 50 candidates per task;
a plan must cover grasp, transfer, placement, and withdrawal.

\paragraph{Results}
Table~\ref{tab:generation-efficiency} reports complete-task plans out of
50 candidates per task and method, evaluated before simulation execution.
Task-guided grasp selection improves planning success from 58\% to 100\%
for Cup-in-Drawer and from 46\% to 90\% for Cup-to-Cabinet.
Failed candidates include grasps that permit pickup but prevent the planner
from finding a collision-free trajectory for subsequent placement or
gripper withdrawal. Task-guided selection accounts for these downstream
requirements when choosing the grasp strategy.

\begin{table}[!t]
\caption{Real-robot policy success rates.}
\label{tab:policy}
\centering
\small
\begin{tabularx}{\columnwidth}{@{}l Y Y Y@{}}
\toprule
Task & \shortstack{Simulation\\demos} & \shortstack{Successes/\\trials} &
\shortstack{Success\\rate (\%)} \\
\midrule
Banana Drawer Storage & 200 & 10/20 & 50 \\
Close Microwave & 100 & 14/20 & 70 \\
Two-Fruit Placement & 200 & 11/20 & 55 \\
\bottomrule
\end{tabularx}
\end{table}

\subsection{Policy Learning and Real-Robot Evaluation}
\label{sec:real-evaluation}

We fine-tune the pretrained $\pi_{0.5}$ vision-language-action
model~\cite{pi2025pi05} for each task evaluated on the real robot, using
only generated simulation demonstrations. Policies receive RGB images,
proprioceptive states, and the shared task instructions. Execution labels
are retained as dataset metadata. We perform full-parameter
fine-tuning with openpi for 15,000 steps, using a batch size of 256 and a
learning rate of $5\times10^{-5}$. The action chunk length
is 10, with all 10 actions executed before the next query.

Table~\ref{tab:policy} reports dataset sizes and complete-task success over
20 real trials per task with layout variations and no human intervention.
Success rates of 50--70\% demonstrate sim-to-real transfer for door pushing,
sequential fruit placement, and object placement followed by drawer
closure. Figure~\ref{fig:platform-rollouts} pairs simulation initial states
with autonomous real executions. The learned policies also exhibit
behavioral diversity: the same Two-Fruit Placement policy successfully
uses both placement orders, and Close Microwave executions use different
door contact positions.

Failures occur when target objects are occluded or when the microwave and
drawer cabinet positions differ substantially from those represented in
the generated demonstrations. These observations indicate sensitivity to
occlusion and spatial layout changes in the deployed policies.

\section{Limitations}
\label{sec:limitations}

The framework depends on the quality of inferred knowledge and the target
scene model. Task ambiguity, occlusion, and motion-recovery errors can lead
to incorrect task conditions, excluding valid alternatives or admitting
inappropriate ones. Geometric and physical validation checks execution
against the inferred conditions but cannot establish whether those
conditions capture the intended task. Errors in target geometry, articulation, and
contact parameters, together with visual and dynamic differences between
simulation and hardware, can degrade generated demonstrations and real-world
execution. Extending the framework to new tasks requires compatible
assets, support for the required interactions, and policy fine-tuning.
Multiple demonstrations and real
execution feedback could improve condition inference and scene-model
accuracy.

\section{Conclusion}
\label{sec:conclusion}

We presented \systemname{}, a framework that uses structured manipulation
knowledge from human videos to guide robot demonstration generation in a
target workspace. To expand behavioral diversity while preserving task requirements,
our VLM-based knowledge extraction and reasoning module distinguishes
required conditions,
demonstration references, and permissible variations within CRAM action
descriptions. To reduce invalid generation attempts, we ground these
descriptions in target-scene entities and geometry to guide candidate
sampling and screening before planning and simulation. The resulting
demonstrations share the inferred task instruction and retain structured
execution labels. Experiments demonstrate additional verified execution
modes beyond a reference-only configuration, improved candidate planning
success through task-guided grasp sampling,
and sim-to-real transfer using $\pi_{0.5}$ fine-tuned on generated
demonstrations.

\bibliographystyle{IEEEtran}
\bibliography{references}

\end{document}